\documentclass[conference]{IEEEtran}
\IEEEoverridecommandlockouts

\usepackage{cite}
\usepackage{amsmath,amssymb,amsfonts}
\usepackage{algorithmic}
\usepackage{graphicx}
\usepackage{textcomp}
\usepackage{xcolor}
\def\BibTeX{{\rm B\kern-.05em{\sc i\kern-.025em b}\kern-.08em
    T\kern-.1667em\lower.7ex\hbox{E}\kern-.125emX}}
\usepackage{multirow}
\usepackage{subcaption}
\usepackage{caption} 
\usepackage{hyperref}
\usepackage{hypcap} 

\usepackage{comment}
\usepackage{comment}
\usepackage{xcolor, color, soul}
\sethlcolor{yellow}
\usepackage{gensymb}
\usepackage{lipsum}
\usepackage{graphicx}
\usepackage{mwe}
\usepackage{booktabs}
\usepackage{balance}

\renewcommand{\baselinestretch}{0.995}

\begin{document}

\title{Evaluating Efficiency and Engagement in Scripted and LLM-Enhanced Human-Robot Interactions} 

\author{\IEEEauthorblockN{Tim Schreiter\textsuperscript{*}}
\IEEEauthorblockA{\textit{PercInS}, 
\textit{Technical University of Munich}\\
Munich, Germany \\
}
\and
\IEEEauthorblockN{Jens V. Rüppel\textsuperscript{*}}
\IEEEauthorblockA{ 
\textit{Chemnitz University of Technology}\\
Chemnitz, Germany \\
}
\and
\IEEEauthorblockN{Rishi Hazra}
\IEEEauthorblockA{\textit{AASS}, 
\textit{\"{O}rebro University}\\
\"{O}rebro, Sweden \\
}
\and
\IEEEauthorblockN{Andrey Rudenko}
\IEEEauthorblockA{\textit{ Corporate Research},
\textit{Robert Bosch GmbH}\\
Stuttgart, Germany 
}
\and
\IEEEauthorblockN{Martin Magnusson}
\IEEEauthorblockA{\textit{AASS}, 
\textit{\"{O}rebro University}\\
\"{O}rebro, Sweden 
}
\and
\IEEEauthorblockN{Achim J. Lilienthal}
\IEEEauthorblockA{\textit{PercInS}, 
\textit{Technical University of Munich}\\
Munich, Germany 
}
\thanks{This work was supported by the European Union's Horizon 2020 research and innovation program under agreement 101017274 (DARKO).}
\thanks{\textsuperscript{*}Authors contributed equally to the work.}
\thanks{Corresponding author \url{tim.schreiter@tum.de}}
}

\maketitle

\begin{abstract}
To achieve natural and intuitive interaction with people, HRI frameworks combine a wide array of methods for human perception, intention communication, human-aware navigation and collaborative action. In practice, when encountering unpredictable behavior of people or unexpected states of the environment, these frameworks may lack the ability to dynamically recognize such states, adapt and recover to resume the interaction. 
Large Language Models (LLMs), owing to their advanced reasoning capabilities and context retention, present a promising solution for enhancing robot adaptability.
This potential, however, may not directly translate to improved interaction metrics. This paper considers a representative interaction with an industrial robot involving approach, instruction, and object manipulation, implemented in two conditions: (1) fully scripted and (2) including LLM-enhanced responses. We use gaze tracking and questionnaires to measure the participants' task efficiency, engagement, and robot perception. The results indicate higher subjective ratings for the LLM condition, but objective metrics show that the scripted condition performs comparably, particularly in efficiency and focus during simple tasks. We also note that the scripted condition may have an edge over LLM-enhanced responses in terms of response latency and energy consumption, especially for trivial and repetitive interactions.
\end{abstract}


\section{Introduction}\label{sec:intro}

The overarching objective of Human-Robot Interaction (HRI) research is the development of robots capable of engaging with humans naturally and intuitively \cite{sheridan2016human, robinson2023robotic}. To reach this objective, researchers develop and combine various methods to perceive and communicate with people, navigate in human environments, and perform collaborative actions. In practice, such systems are typically implemented with partially scripted scenarios or pre-programmed behaviors \cite{sheridan2016human, tsarouchi2016human}. While these approaches work well in controlled settings, they often lack the ability to adapt to inherently dynamic and unpredictable human behavior. Robots such as the NAO, commonly deployed as transmitters of social cues \cite{masson2017nao, schreiter2023advantages} by researchers, often utilize such scripted behaviors \cite{amirova21}. The lack of flexibility becomes particularly problematic when testing new behaviors in user studies, as minor deviations from the expected protocol by participants can lead to invalid experiment states or unsuccessful outcomes. Another method deployed to compensate for technical limitations is the Wizard of Oz method \cite{riek2012wizard}. However, this approach does not only still suffer from limited contextual inputs and intervention capabilities, it also ultimately replaces human-robot interaction with human-human interaction, mediated by a robot \cite{rudenko2024child}, 
always at the risk of failing to create the illusion of autonomous interaction.

LLMs represent a promising opportunity to facilitate flexible and context-aware responses to user inputs, potentially adapting and recovering from unexpected interaction states. However, augmenting scripted interactions with LLM-enhanced actions is not trivial due to the complexity and cost of integrating these models into existing systems \cite{xia2024applying}. The integration process requires substantial computational infrastructure, specialized expertise in model deployment, and careful consideration of the system architecture. Additionally, researchers must address conversational coherence and manage latency in real-time interactions \cite{callie2024understand}. Moreover, the resource-intensive nature of LLMs, including their computational demands and energy consumption \cite{de2023growing}, necessitates a careful cost-benefit analysis when considering an implementation in robotic systems.

This paper compares a scripted human-robot interaction with one in which an LLM is used to generate responses dynamically for a collaborative pick-and-place task usually found in industrial applications \cite{makris2024seamless}. We consider the participants' visual attention as an input to the LLM framework and also use it to quantify their perception of the robot. Additionally, we measure the participants' task performance
and use questionnaires to assess their subjective ratings of the robot. By analyzing participants' gaze patterns, we aim to understand if and how attention allocation differs between the two conditions, mainly focusing on key interaction moments and robot communication behaviors. Our experimental design employs a within-subjects approach, where participants experience both interaction conditions in random order, allowing for direct comparison of behavioral and perceptual differences. Our results offer valuable insights into the potential of LLMs to enhance classical HRI methods by making interactions more adaptive and robust against unexpected human behavior. We also emphasize the importance of selecting interaction approaches that match the specific demands and complexity of the task, taking into account factors such as technical implementation challenges, energy consumption, and the trade-off between adaptability and predictability in the user experience.

\begin{figure}[t]
\vspace{0.2cm}
\centering
\includegraphics[width=\columnwidth]{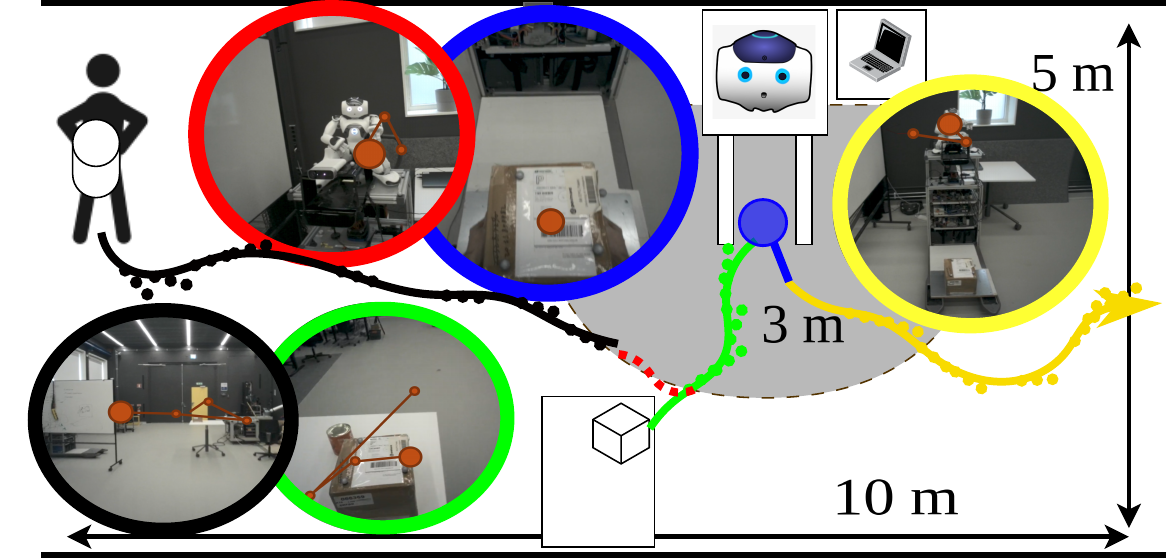}
\caption{\small{Setup for interacting with a NAO robot mounted on a forklift. 
The participant begins at the corridor entrance, the robot is located behind a wall. Interaction steps: \textbf{Black:} Carrying and placing a tin can; \textbf{Red:} Receiving instructions from the NAO robot; \textbf{Green, Blue:} Box delivery; \textbf{Yellow:} Interaction conclusion. Insets show egocentric eye-tracking with gaze sequences overlaid in \textbf{brown}.}}
\label{abb:expsetup}
\end{figure}

\section{Methods}

This paper considers a representative interaction with an industrial forklift robot mediated by an anthropomorphic NAO robot~ \cite{schreiter2023advantages}. During the interaction, the robot asks the participant to approach, manipulate objects, and disengage. The interaction is implemented in two conditions: one where the steps are scripted and timed, and another where an LLM backbone is used to generate flexible actions, adapting to the interaction status. We aim to quantify the difference in robot perception and task execution between those two conditions. To that end we analyze gaze tracking and questionnaire responses, and measure energy consumption in both conditions by computing power across network nodes and API service calls.

\subsection{Experiment Design}\label{exp_design}

\begin{figure}[t]
\vspace{0.2cm}
\centering
\includegraphics[width=8cm]{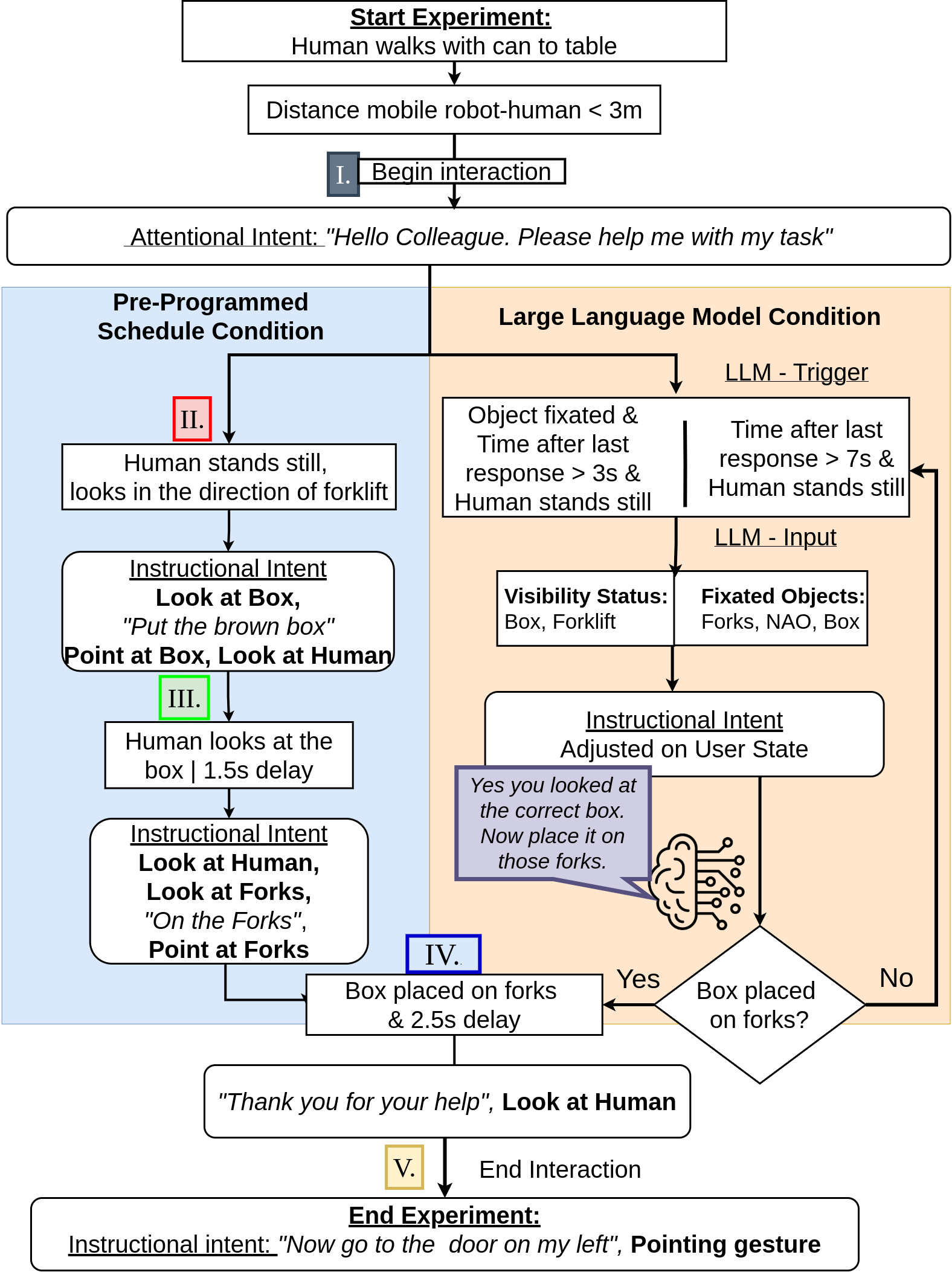}
\caption{\small{Sequence plan of the experiment. The robot is either controlled by an LLM (in \textbf{blue}) or a pre-programmed schedule (PPS, in \textbf{orange}). Roman numbers I--IV denote the key interaction steps, color coded according to Figure \ref{abb:expsetup}.
}}
\label{abb:pap}
\end{figure}

The study was conducted in a 10~m x 5~m corridor. Figure~ \ref{abb:expsetup} illustrates a sample interaction including sample cutouts of the recorded gaze overlays to show what participants observed during the interaction. Participants began the experiment at one end of the corridor, holding a tin can with both hands. Their initial task (given by the experimenter) was to approach a table on their right and place the tin can there. During this stage, the robot they would later interact with was occluded and not visible to them. After the participant had placed the tin can on the table and approached the forklift ($\leq$ 3 m),
the robot initiated interaction with a predefined phrase: ``Hello, colleague. Please help me with my task.''

The subsequent interaction followed a sequence plan implemented in two conditions (pre-programmed schedule, or PPS, and LLM-enhanced), as depicted in Figure \ref{abb:pap}. The key difference between the two conditions was the interaction control mechanism. In the PPS condition, the robot's behavior followed a fixed schedule based on the design in \cite{schreiter2023advantages}, providing deterministic and predictable cues without adapting to the participant's specific behavior. In contrast, LLM-enhanced responses utilized an interaction interface incorporating multiple visual and gaze attention cues and LLM reasoning abilities. It considered possibly visible objects in the participants' field of view and the objects they fixated on, providing action suggestions based on the egocentric view of the participant and the task description. The LLM received these inputs in text form, i.e., ``These objects are detected in the scene'' and ``User fixated on these objects''.
In both conditions, the NAO robot, acting as a mediator for the forklift \cite{schreiter2023advantages}, employed multimodal communication throughout the interaction, including speech, referential gestures (pointing) and gaze behaviors (eye contact and head movements) to track and guide the participants. 

In the PPS condition (blue section in Figure \ref{abb:pap}), Step II began once the participant’s walking speed fell below 0.3 m/s. The robot instructed them to pick up a box near the table and place it on the forklift. When participants focused on the box (either immediately or after a 1.5 s delay), the robot specified where to place it. The robot used referential gestures and gaze throughout this process to improve clarity. If participants did not place the box on the forks, the robot repeated its directive until they placed it, triggering Step IV after 2.5 s.

In the LLM condition (orange section in Figure \ref{abb:pap}), a custom framework \cite{schreiter2024bidirectional} combining ``GPT 4o-mini'', mobile eye-tracking (Tobii Glasses 3) with YOLOv8-based object detection to guide interactions in real-time. Rather than using fixed transitions, it adapted to a participant’s gaze and locomotion. For example, a stationary participant (moving at $v \leq 0.3$ m/s) triggered a step by fixating on an object (such as a forklift or box) for more than three seconds without receiving any recent instructions. After seven seconds of inactivity, the robot prompted further action to avoid prolonged idling. Gaze data, captured as pixel coordinates and mapped to object bounding boxes, identified fixations lasting at least 30 ms. This data was sent as a Python dictionary to the LLM to generate real-time prompts. Once the participant placed the box on the forklift (Step IV), both conditions aligned again. Using OpenAI’s ``Chat Completions,'' the LLM maintained context and used Chain-of-Thought Reasoning \cite{wei2022chain} to produce code for NAO’s pointing, gaze, and speech. The robot then thanked the participants and instructed them to disengage.

\subsection{Participants}\label{subsec:samples}

We recruited 15 participants for the experiment, aged 20 to 65 years (M = 30.1, SD = 10.8). All participants are fluent in English and identify as female (7/15), male (7/15), or preferred not to answer the question (1/15). Written informed consent was obtained from all participants prior to the experiment, following consultation with local authorities. Participants experienced both conditions in random order to counterbalance possible learning effects. Participants also received instructions on proper forklift interaction techniques, as they were not expected to have prior experience with such equipment.

\subsection{Data Collection}
Eye-tracking data was obtained from the Tobii Pro Glasses 3. We applied the standard Tobii I-VT gaze filter with a classification threshold of $100^\circ/$s and used Tobii's software for evaluation. We measured visual engagement through fixation duration (sustained attention on specific areas), saccade velocity- and amplitude (attention shifts and scanning behavior), pupil diameter (cognitive activity), and color-coded heatmaps displaying fixation density. All metrics were recorded separately for task elements and robot interaction. To correctly resolve the participants and robot position during the experiment, the eye-tracking data is supported by a motion capture system. Passive IR markers were attached to the robot and to the helmets that participants wore to resolve the orientation and position in 6D.

In addition to eye-tracking metrics, we measured subjective perceptions of the robot using questionnaires. We employed the ``Trust in Industrial Human-Robot Collaboration'' scale \cite{charalambous2016development} to assess how each interaction condition affects user trust. We applied changes regarding the gripper-related items as described by the authors of \cite{schreiter2022effect}. Additionally, we used Bartneck’s ``Godspeed Questionnaire'' \cite{bartneck2009measurement} to better understand participants’ perceptions of the robot system.

We measured time allocation and visual attention in both conditions separately in the task execution and robot interaction phases. After marking distinct phases of task-related actions and robot communication events, participants' durations were extracted from the eye-tracking software. Additionally, direct power measurements quantified energy consumption across both interaction modalities. A single-machine power monitor was utilized for the PPS condition. At the same time, we incorporated the LLM condition's measurements as a sum of computing power across network nodes and API service calls. The analysis factored in both operational power consumption and the training energy costs of the LLM.

\section{Results}\label{sec:result}

The results of the Trust scale show a slightly higher median rating for the LLM condition (44) than for the PPS condition (42). A one-way ANOVA ($F=1.05$, $p=0.32$) showed the result was not statistically significant. The Godspeed questionnaire assessed participants' perceptions of the robot in both conditions. As the data was not normally distributed (confirmed by Shapiro-Wilk tests), we used Mann-Whitney U tests, which showed minor, non-significant differences ($p>0.05$) across subscales: Anthropomorphism (Medians of 13/25 for LLM; 12/25 for PPS), Animacy (20/30 for LLM; 19/30 for PPS), Likeability (both 20/25), Perceived Intelligence (17/25 for LLM; 16/25 for PPS), and Safety (both 11/15).

\begin{figure*}
    \centering
    \begin{subfigure}[b]{0.23\textwidth}
        \centering
        \includegraphics[width=\textwidth, height=2cm]{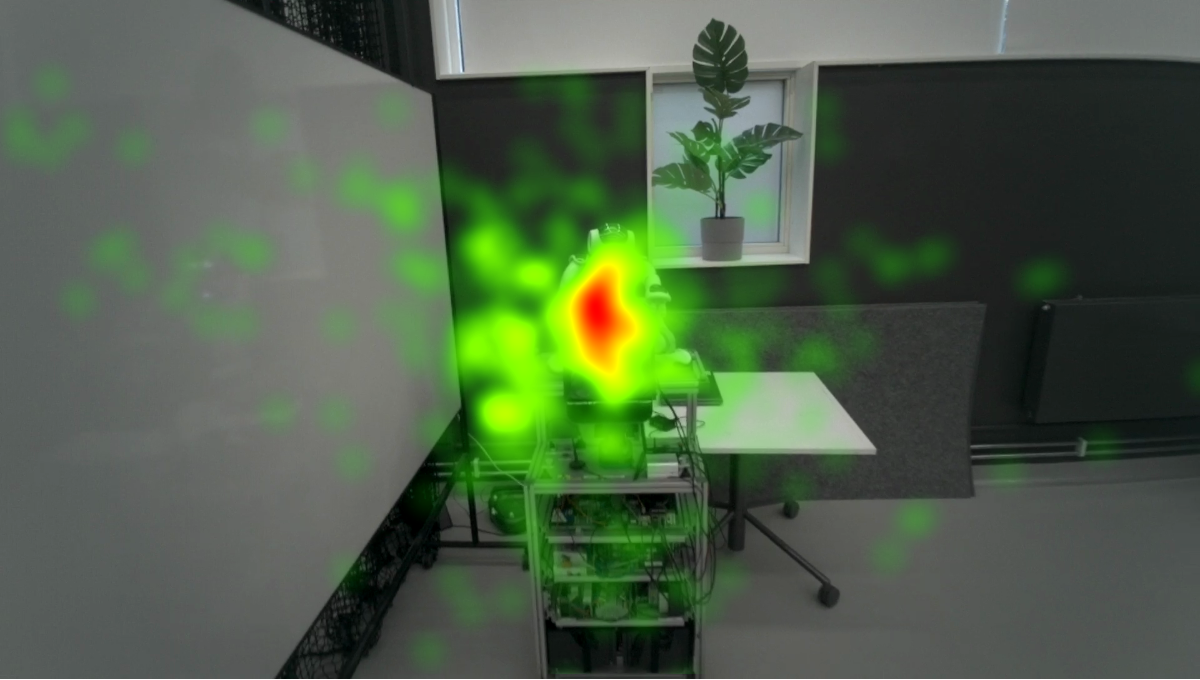}
        \caption[LLM Interaction]%
        {{\small LLM-enhanced interaction}}    
        \label{fig:llm_interaction}
    \end{subfigure}
    \hfill
    \begin{subfigure}[b]{0.23\textwidth}  
        \centering 
        \includegraphics[width=\textwidth, height=2cm]{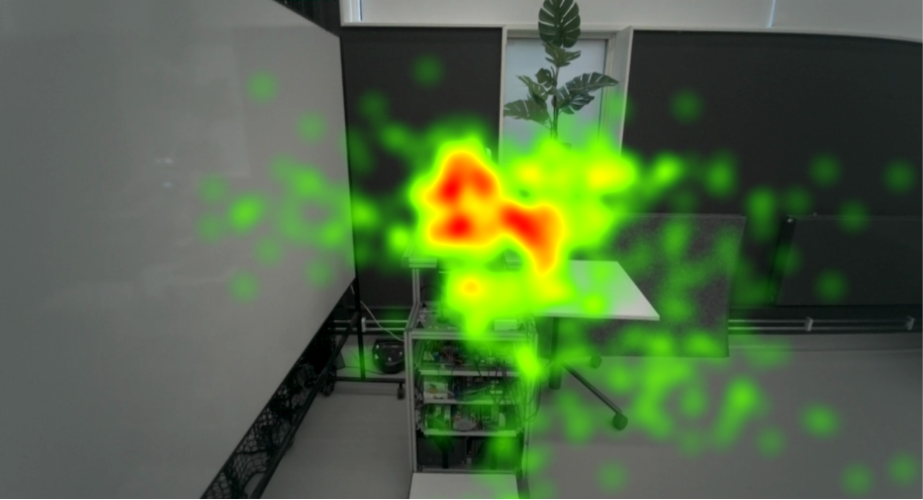}
        \caption[PPS Interaction]%
        {{\small PPS-controlled interaction}}    
        \label{fig:pps_interaction}
    \end{subfigure}
    \hfill
    \begin{subfigure}[b]{0.23\textwidth}   
        \centering 
        \includegraphics[width=\textwidth, height=2cm]{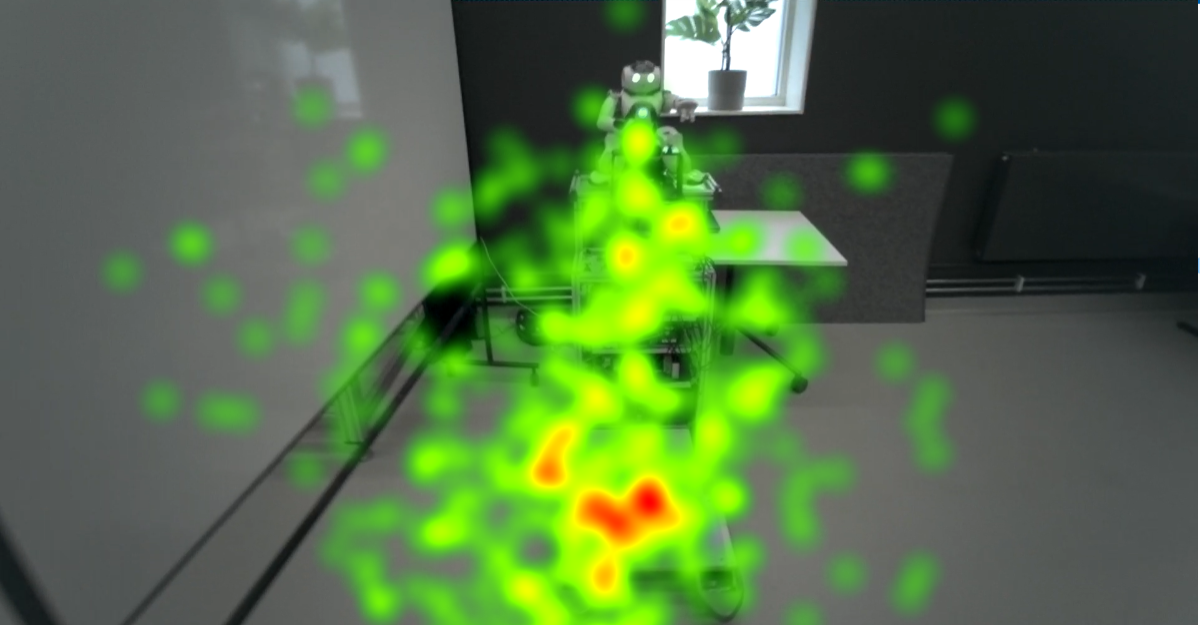}
        \caption[LLM Task Execution]%
        {{\small LLM-enhanced task}}    
        \label{fig:llm_task}
    \end{subfigure}
    \hfill
    \begin{subfigure}[b]{0.23\textwidth}   
        \centering 
        \includegraphics[width=\textwidth, height=2cm]{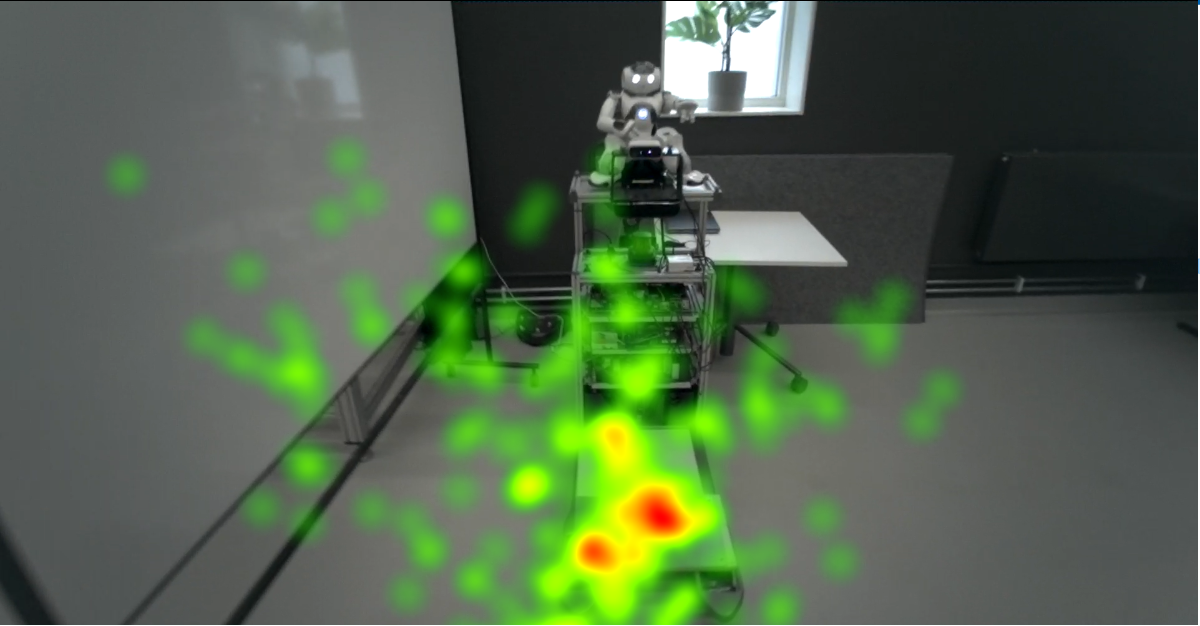}
        \caption[PPS Task Execution]%
        {{\small PPS-guided task}}    
        \label{fig:pps_task}
    \end{subfigure}
    \caption[Comparison of LLM and PPS Interactions]
    {\small Fixation heatmaps in the interaction and task execution phases for the LLM and PPS conditions. Subfigures (a) and (b) highlight differences in attention during robot interaction. Subfigures (c) and (d) highlight differences in visual attention during task execution.} 
    \label{fig:critical_parameters}
\end{figure*}

The heatmaps in Figure \ref{fig:critical_parameters} show distinct visual attention patterns in both conditions during the interaction and task execution phases. In the LLM condition, fixations are almost exclusively on the robot's upper body during the interaction and are widely spread across the robotic forklift during the task. In contrast, in the PPS condition fixation patterns are more dispersed with broader visual exploration during the interaction phase but more focused during task execution.

\begin{table}[t]
\centering
\scriptsize
\caption{\small{Gaze metrics during task execution and interaction with the robot in the LLM and PPS conditions.}}
\begin{tabular}{lccc}
\toprule
\textbf{Metric (Phase)} & \textbf{PPS} & \textbf{LLM} \\ \midrule
Interval Duration [s] (Task Execution)*        & 10.1 ± 3.48   & 11.42 ± 10.6  \\
Interval Duration [s] (Interaction)    & 10.43 ± 4.23   & 12.39 ± 4.00  \\
Fixation Duration [s] (Task Execution)*        & 70.08 ± 24.31  & 79.08 ± 74.73 \\
Fixation Duration [s] (Interaction)    & 67.49 ± 28.09  & 82.38 ± 27.44 \\
Pupil Diameter [mm] (Task Execution)           & 3.66 ± 0.58    & 3.66 ± 0.59   \\
Pupil Diameter [mm] (Interaction)       & 3.57 ± 0.59    & 3.57 ± 0.53   \\
Saccade Velocity [$\degree/s$] (Task Execution)*         & 434.05 ± 89.49 & 420.03 ± 128.92 \\
Saccade Velocity [$\degree/s$] (Interaction)     & 332.29 ± 75.63 & 380.05 ± 79.09 \\
Saccade Amplitude [$\degree$] (Task Execution)*        & 130.64 ± 83.66 & 126.47 ± 98.09 \\
Saccade Amplitude [$\degree$] (Interaction)**    & 63.59 ± 33.32  & 82.48 ± 39.11 \\ \bottomrule
\end{tabular}
\footnotesize \textbf{Note:} *$p<0.05$; **$p<0.01$
\label{table:combined_comparison_metrics}
\vspace{-0.3cm}
\end{table}

In addition to the heatmaps, other gaze tracking metrics highlighted engagement differences between the two conditions as shown in Table \ref{table:combined_comparison_metrics}. The LLM condition showed a longer sum of all fixation durations, indicating sustained attention to task elements and the robot. The LLM condition exhibited slightly lower peak saccade velocities and amplitudes during task execution. The PPS condition recorded shorter fixation durations, with slightly higher saccade velocities and amplitudes. The pupil diameter, as a measure of cognitive load \cite{Kahneman1966PupilDA}, was consistent across both conditions, indicating a similar cognitive load. The objective measures show differences in time allocation between the LLM-enhanced and PPS-controlled conditions. The participants spent 31.65\% (LLM), 15.74\% (PPS) of their time on task execution, and 52.95\% (LLM), 47.86\% (PPS) on robot interaction. The remainder of the time was spent on approach, waiting, and disengagement times 15.40\% (LLM), 36.40\% (PPS). The lower task time ratio in the PPS condition indicates less engagement with task elements, potentially due to the predictability of the PPS.

The energy consumption of robot interaction with the user-centric framework \cite{schreiter2024bidirectional} varied between the two conditions as expected.
In the PPS condition (single-machine computation), the energy consumption per interaction was found to be 506~Wh. The LLM condition exhibited a base consumption of 491 Wh per interaction. We supplement this by accounting for the energy consumed during model training and inference. Referring to a recent study \cite{de2023growing} for GPT-3's total training energy (1,287,000 kWh) and dividing it by its estimated annual queries (71.2B = 195M daily × 365), we calculated an additional 18 Wh per interaction. As for inference in \cite{de2023growing}, the daily energy consumption is approximated with 564 MWh (based on the number of GPUs used), which would add 3 Wh (564MWh * 365 days = 205860 MWh / 71.2B interactions $\approx$ 3 Wh) per interaction resulting in a total of 21Wh per API call. With 2-4 API queries per participant, the total energy consumption of the LLM condition ranged from 538 to 580 Wh per interaction. Although our study employed GPT-4o-mini, we used GPT-3 data as a proxy due to the unavailability of specific energy metrics for the newer model.

\section{Discussion}\label{sec:discuss}

Our participants rated the LLM condition marginally higher (but not statistically significant) in trust and anthropomorphism.
This may indicate that users feel more comfortable 
with robots offering dynamic, context-aware responses, which is beneficial in industrial settings where trust impacts productivity and safety. Still, the marginal difference indicates that PPS remains a viable, simpler option for the examined task.

The analysis of eye-tracking data revealed distinct visual attention patterns for both conditions. We observed longer fixation durations in the LLM condition, indicating a sustained engagement with both the robot and task elements due to the adaptive, context-sensitive cues produced by the LLM. The heatmaps for the LLM condition showed that participants concentrated on the NAO's upper body during the interaction phase while distributing task-related fixations across the forklift. This sustained focus may benefit scenarios requiring frequent interaction or focused monitoring, enhancing long-term interaction quality. On the other hand, the PPS condition exhibited shorter fixations and faster saccades, with more dispersed gaze patterns during interaction and focused attention during task execution. These differences suggest that while the LLM adaptability supports complex, dynamic tasks, the predictability of PPS and focused engagement may be better suited for repetitive, efficiency-driven applications.

Our analysis revealed distinct operational characteristics of the two conditions. The PPS condition delivered deterministic interactions independent of network connectivity, with commands structured into discrete steps (e.g., ``Put the brown box''; ``On the forks''). This approach minimized redundancy and reduced response latency, which would be particularly effective for repetitive tasks. The LLM condition exhibited contextual adaptation, modifying instructions based on the participant's behavior and environment state. While its interactive feedback (e.g., ``It seems you are looking in the right direction'') enhanced engagement, it occasionally produced instruction redundancies. Therefore, this condition might better suit complex tasks requiring dynamic adaptation.

Enhancing human-robot interaction with LLMs presents development and deployment challenges. Dependencies on a stable network connection and OpenAI API impact system robustness, while the output variability of the LLM requires prompt engineering to maintain interaction consistency. Prompting the LLM to respond with code generation produces more structured outputs than natural language responses. However, additional parsing mechanisms were required to ensure efficient interactions. The observed response latency of the API of 2.5~s constrained real-time interaction capabilities. Future implementations could leverage open-source models (e.g., Llama 40B or Mixtral models) for local deployment, mitigating the latency issue and reducing operational costs, given an adequate computational infrastructure.

Our findings highlight the importance of aligning interaction modalities with task requirements. The PPS is suitable for straightforward tasks due to its predictability, efficiency, and low energy demands, especially for systems with limited computational resources. In contrast, an LLM's adaptability and dynamic guidance offer advantages in complex, interactive scenarios but are subject to higher implementation complexity and operational demands.



\bibliographystyle{IEEEtran}
\balance
\bibliography{IEEEabrv,references}

\end{document}